\documentclass[11pt,a4paper]{article}
\usepackage[hyperref]{acl2021}
\usepackage{times}
\usepackage{graphicx}
\usepackage{multirow}
\usepackage{amsmath}
\usepackage{latexsym}

\usepackage{microtype}

\aclfinalcopy % Uncomment this line for the final submission
\usepackage[utf8]{inputenc}

\newcommand*{\enja}{EN$\rightarrow$JA }
\newcommand*{\jaen}{JA$\rightarrow$EN }

\title{Revisiting Context Choices for Context-aware Machine Translation}

\author{Mat{\=\i}ss Rikters and Toshiaki Nakazawa \\
  The University of Tokyo \\
  7-3-1 Hongo, Bunkyo-ku, Tokyo, Japan \\
  \texttt{\{matiss, nakazawa\}@logos.t.u-tokyo.ac.jp} \\}

\date{}

\begin{document}
\maketitle
\begin{abstract}
One of the most popular methods for context-aware machine translation (MT) is to use separate encoders for the source sentence and context as multiple sources for one target sentence. Recent work has cast doubt on whether these models actually learn useful signals from the context or are improvements in automatic evaluation metrics just a side-effect. We show that multi-source transformer models improve MT over standard transformer-base models even with empty lines provided as context, but the translation quality improves significantly (1.51 - 2.65 BLEU) when a sufficient amount of correct context is provided. We also show that even though randomly shuffling in-domain context can also improve over baselines, the correct context further improves translation quality and random out-of-domain context further degrades it.
\end{abstract}

\section{Introduction}

There are two main approaches for incorporating context in neural machine translation (NMT) along with several others, which have not been adapted as widely \cite{lopes-etal-2020-document}. The most common methods are: 1) data concatenation without changing the model architecture \cite{tiedemann-scherrer-2017-neural}, which can even be extended to full documents on both source and target \cite{junczys-dowmunt-2019-microsoft}; 2) training models with multiple separate encoders for main sentences and context \cite{zhang-etal-2018-improving}; and 3) other methods, such as cache-based \cite{tu-etal-2018-learning}, hierarchical attention \cite{miculicich-etal-2018-document,maruf-etal-2019-selective}. 
The first approach faces the challenge of encoding longer than usual inputs and separating where context ends and content begins. In this work, we focus on the multi-encoder approach and aim to explore three main research questions: 1) is there an optimal amount of previous context sentences; 2) how is translation quality affected by training models using random in-domain context sentences as opposed to random out-of-domain context sentences; and 3) how will translation quality change when using empty lines as context.

Recently there have been several studies \cite{kim-etal-2019-document,li-etal-2020-multi-encoder,jwalapuram2020benchmark} that attribute the success of context-aware NMT to regularisation and noise generated by the additional encoders rather than the actual context. While we are confident that the model alone plays a substantial role in this, we do believe that the correct data matters and even speculate that the second encoder may act as a sort of domain adaptation mechanism on the context data.

We wish to direct our research towards the Japanese-English language pair, as it is very common in the Japanese language to omit pronouns and obvious arguments to verbs which can be inferred from context.

\section{Related Work}

Multi-source and other multi-encoder models have been widely used in several language processing tasks, such as automatic post-editing \cite{junczys-dowmunt-grundkiewicz-2018-ms,shin-lee-2018-multi}, speech recognition \cite{Zhou2020}, multilingual MT \cite{zoph-knight-2016-multi} and multimodal MT \cite{yao-wan-2020-multimodal}. There are also many studies on using multi-encoder models for context-aware MT \cite{jean2017does,zhang-etal-2018-improving} with various degrees of success.

\begin{figure}[t]
  \includegraphics[width=\linewidth]{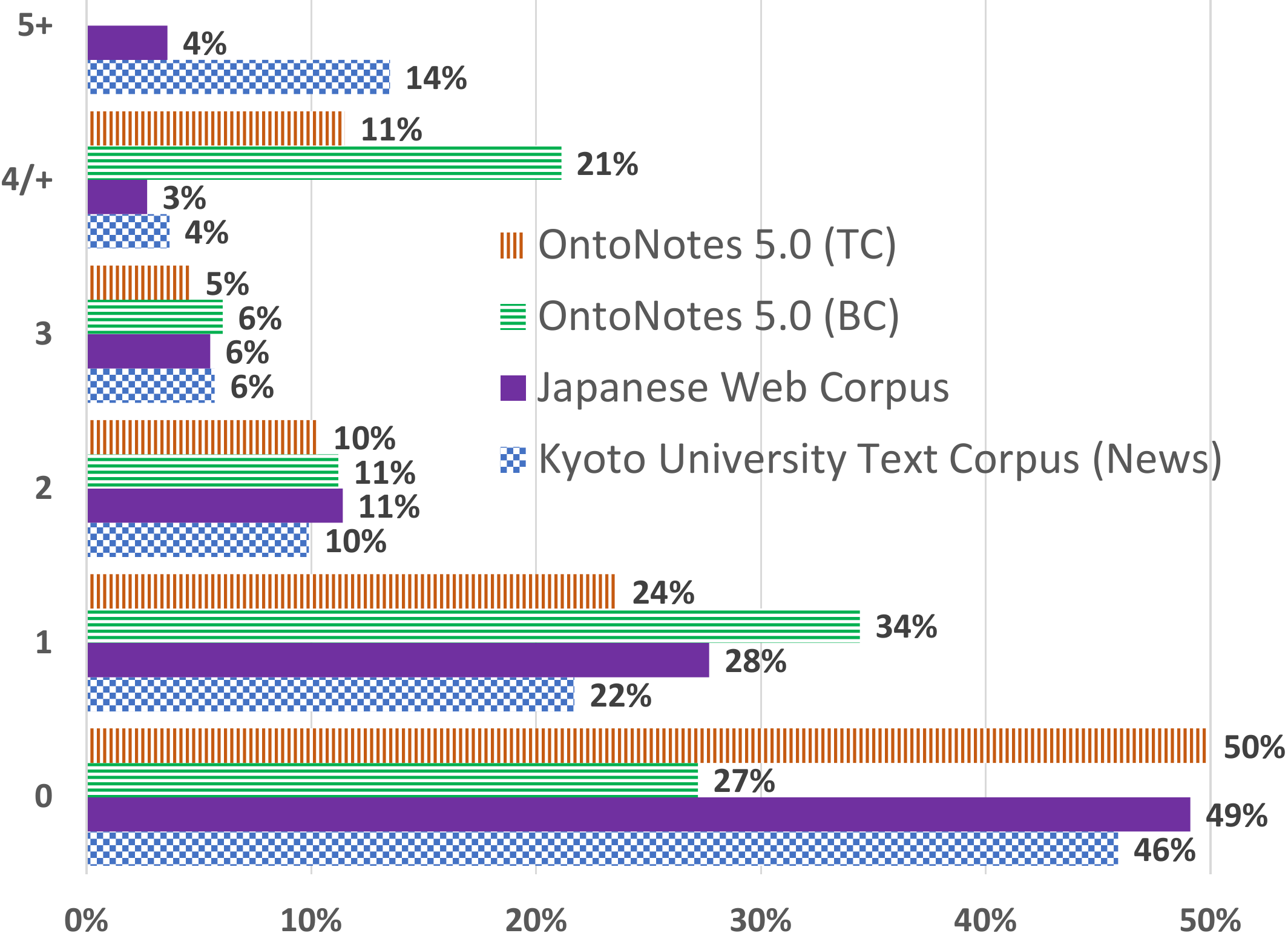}
  \caption{Distribution of the position of the antecedent from the current sentence in Japanese and English. 0 means the antecedent of a anaphora is in the same sentence, 1 means the previous sentence and so on. For Japanese corpora (Kyoto University Text Corpus and Japanese Web Corpus), the antecedent of the omitted element (zero-anaphora) are investigated. For English corpora (OntoNotes 5.0 Broadcast Conversation and Telephone Conversations), the antecedent of all coreference relations are investigated.}
  \label{fig:antecedent-dist}
\end{figure}

Most related work focuses on either only 1-2 previous/next sentences as context \cite{tiedemann-scherrer-2017-neural,voita-etal-2018-context} or training on full documents \cite{junczys-dowmunt-2019-microsoft,mace_valentin_2019_3525020} with nothing in between. From our preliminary investigation of English corpus (OntoNotes 5.0) and findings of the previous work \citet{Masatsugu-Hangyo2014}, more than 20\% of the antecedents of anaphoras appear more than two sentences before the current sentence where the anaphoras appear. The detailed distribution of the position of the antecedents are shown in Figure \ref{fig:antecedent-dist}. This indicates that the existing models which consider only 1-2 previous sentences are not sufficient. In this paper, we explore variations of using 0-4 previous sentences as context.

\citet{li-etal-2020-multi-encoder} experimented with training models using a fixed sentence or randomly sampled words from the vocabulary as context and compared the results to using the actual previous sentence as context. They found that the model still improves over the baseline transformer even with incorrect context and in some cases even outperforms models trained with correct context.

\citet{kim-etal-2019-document} explore how removing specific parts of the context impacts multi-encoder MT performance. 
They find that actual utilisation of document-level context is rarely interpretable, but filtering out stop-words and most frequent words from the context or keeping only named entities or specific parts of speech (POS) does not strongly impact translation quality. They also experiment with using longer context of up to 20 sentences and show that performance drops with more than 1-2 sentence context when using full sentences, but is more stable and even improves when retaining only specific POS in the context.

\citet{stojanovski2020addressing} train multi-domain models for translating from English into German using the concatenation approach with 1, 5 and 10 context sentences and separate embeddings to specify the domain of each sentence. They find that the best results are from either the 5 or 10 context sentence model, depending on the domain of the evaluation data.
They also perform an ablation experiment by providing context from a different domain at evaluation time for a model trained on the correct context. 

We hypothesise that it is actually not the random or fixed sentences provided as context that improve multi-encoder MT output over the baseline, but rather the larger model architecture. 
To prove this, we extend these experiments by training models with empty lines as context and show that that alone is enough to outperform the baseline transformer model.

We also believe that using random tokens from the same training corpus as context improves the final translation by essentially performing as domain adaptation. To verify this claim, we train separate models using 1) randomly sampled sentences from the same corpus and 2) randomly sampled sentences from a completely different corpus as context. We find clear differences between the two randoms as well as a difference between the best random and best actual context model.

\section{Multi-source Transformer Model}

There are several different ways to implement multi-source encoder models for MT like concatenating outputs from multiple encoders \cite{pal-etal-2018-transformer} or averaging them. For our experiments we follow the approach that \citet{junczys-dowmunt-grundkiewicz-2018-ms} used for automatic post-editing, where the original transformer \cite{NIPS2017_3f5ee243} is supplemented by a second encoder and a second multi-head attention block is stacked above the previous multi-head attention block. 

We consider two main configurations for training our models, which differ only by the data that is provided to the second encoder as context. The first is \textit{n-context}, where \textit{n} is 0-4 specifying the maximum number\footnote{Note that even if \textit{n} is 4 the first sentence of each document will always have 0 context, the second will have 1 and so on.} of previous context sentences for each content sentence. For the second configuration we chose to use an \textit{n} of 3 (due to good performance in the first configuration and too few antecedents being further as shown in Figure \ref{fig:antecedent-dist}) and train models with random in-domain context (\textit{3-random-ind}) and random out-of-domain context (\textit{3-random-ood}).

\section{Experiments}

We experiment with training EN$\leftrightarrow$JA models with several slightly differing configurations. We used the document-aligned Japanese-English conversation corpus \cite{rikters-EtAl:2020:WMT} as training data, which contains about 220k parallel sentences from about 3k documents. We used the development and evaluation data from the corpus, each containing about 2k sentences from 69 documents, for development and evaluation of our models.

For pre-processing we used only Sentencepiece \cite{kudo-richardson-2018-sentencepiece} to create a shared vocabulary 16k tokens. We did not perform other tokenisation or truecasing for the training data. We used Mecab \cite{kudo2006mecab} to tokenise the Japanese side of the evaluation data, which we used only for scoring. The English side remained as-is.
% The parameter count was about 53M for the baseline transformers and 78M for the multi-source models.
We use Marian \cite{junczys-dowmunt-etal-2018-marian} to train transformer-base models as baselines and seven different configurations of multi-source transformer models using up to 4 previous sentences as context, up to 3 random sentences from the same training data as context, and up to 3 random sentences from a different corpus \cite{morishita-etal-2020-jparacrawl} as context. In all experiments with more than one context sentence the context sentences were provided in a single file to the additional encoder divided by the tabulation symbol.
Each model was trained using three random seeds (347155, 42, 9457) on two TITAN Xp GPUs until convergence (loss not improving for 7 checkpoints) with training time of about one day per model.
We use the SacreBLEU\footnote{Version string: BLEU+case.mixed+numrefs.1+smooth.\\exp+tok.13a+version.1.2.21} tool \cite{post-2018-call} to evaluate automatic translations and calculate BLEU \cite{papineni-etal-2002-bleu}, NIST \cite{10.5555/1289189.1289273} and ChrF \cite{popovic-2015-chrf} scores.

\begin{table}[t]
\small
\centering
\begin{tabular}{|lc|c|c|}
\hline   
\multicolumn{1}{|l|}{\multirow{2}{*}{\textbf{Configuration}}} & \multicolumn{1}{c|}{\textbf{BLEU}} & \multicolumn{1}{c|}{\textbf{NIST}} & \multicolumn{1}{c|}{\textbf{ChrF2}} \\ \cline{2-4}
                                    & \multicolumn{3}{|c|}{\textbf{\jaen}}                                                                              \\ \hline   
\multicolumn{1}{|l|}{baseline}      & 12.35 ± 0.77                         & 3.48 ± 0.11                          & 40.88                               \\ 
\multicolumn{1}{|l|}{0-context}     & 15.31 ± 0.85                         & 4.02 ± 0.14                          & 44.74                               \\ 
\multicolumn{1}{|l|}{1-context}     & 17.29 ± 0.87                         & 4.39 ± 0.14                          & 47.12                               \\ 
\multicolumn{1}{|l|}{2-context}     & 17.34 ± 0.88                         & 4.43 ± 0.14                          & 47.33                               \\ 
\multicolumn{1}{|l|}{3-context}     & 17.96 ± 1.02                         & 4.51 ± 0.14                          & 47.73                               \\ 
\multicolumn{1}{|l|}{4-context}     & 17.14 ± 0.92                         & 4.36 ± 0.14                          & 46.83                               \\
\multicolumn{1}{|l|}{3-random-ind}  & 17.65 ± 0.91                         & 4.52 ± 0.13                          & 47.67                               \\
\multicolumn{1}{|l|}{3-random-ood}  & 16.56 ± 0.92                         & 4.28 ± 0.14                          & 46.45                               \\ \hline
\multicolumn{1}{|l|}{WMT20}        & 16.29                         &      4.33                                & 45.54                               \\ 
\multicolumn{1}{|l|}{WMT20+}        & 18.44                         &     4.81                                 & 48.12                               \\ \hline
                                    & \multicolumn{3}{|c|}{\textbf{\enja}}                                                                              \\ \hline
\multicolumn{1}{|l|}{baseline}      & 11.86 ± 0.71                         & 3.87 ± 0.10                          & 29.68                               \\ 
\multicolumn{1}{|l|}{0-context}     & 14.00 ± 0.76                         & 4.17 ± 0.10                          & 32.21                               \\
\multicolumn{1}{|l|}{1-context}     & 14.93 ± 0.78                         & 4.36 ± 0.10                          & 33.44                               \\ 
\multicolumn{1}{|l|}{2-context}     & 15.51 ± 0.81                         & 4.45 ± 0.10                          & 34.30                               \\ 
\multicolumn{1}{|l|}{3-context}     & 15.26 ± 0.82                         & 4.42 ± 0.11                          & 33.92                               \\ 
\multicolumn{1}{|l|}{4-context}     & 15.19 ± 0.79                         & 4.36 ± 0.11                          & 33.68                               \\ 
\multicolumn{1}{|l|}{3-random-ind}  & 15.18 ± 0.78                         & 4.36 ± 0.10                          & 33.86                               \\ 
\multicolumn{1}{|l|}{3-random-ood}  & 14.43 ± 0.80                         & 4.26 ± 0.10                          & 32.85                               \\ \hline
\multicolumn{1}{|l|}{WMT20}        & 12.99                         &   3.98                                   & 31.09                               \\ 
\multicolumn{1}{|l|}{WMT20+}        & 15.33                         &   4.40                                   & 33.97                               \\ \hline
\end{tabular}
\caption{Automatic evaluation results.}
\label{tab:automatic-result-table}
\end{table}

\begin{figure}[t]
  \includegraphics[width=\linewidth]{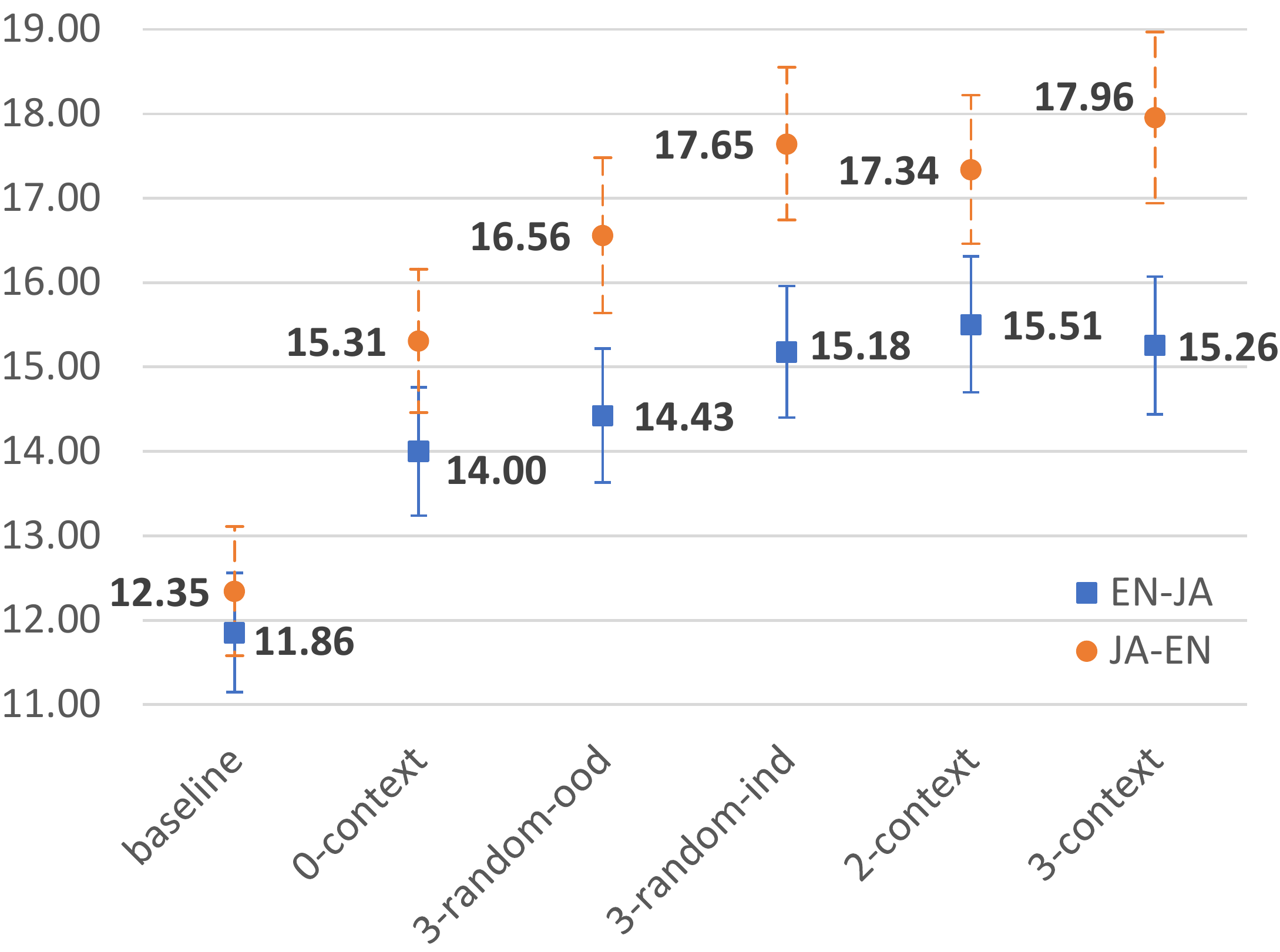}
  \caption{Best EN$\leftrightarrow$JA results compared to the baseline and 0 context, random out-of-domain (ood) context, and random in-domain (ind) context.}
  \label{fig:bootstrap-bleu-figure}
\end{figure}

Experiment results are summarised in Table \ref{tab:automatic-result-table} and the most distinctive results of BLEU scores are visualised in Figure \ref{fig:bootstrap-bleu-figure}. We apply paired bootstrap resampling \cite{koehn-2004-statistical} to calculate significance intervals of BLEU and NIST scores. Here we see that all results significantly outperform the baseline models, even the \textit{0-context} and random context ones. However, if we consider models with \textit{0-context} as our true baseline, then models with out-of-domain random context are within the margin of error while all \jaen models are with in-domain context score significantly higher than that. For \enja there is a slight overlap of 0.06 BLEU in the error intervals between the \textit{0-context} model and the highest scoring model which used 2 context sentences, but according to the NIST score there is significant difference. We further verify the significance of this difference in the human evaluation section.

The results also show that there are differences in automatic evaluation scores between all models trained using 1-4 or random 3 sentence actual context, but they are within the margin of error of each other. Nevertheless, we can see that in both translation directions using 3 actual context sentences is slightly better than 3 random in-domain context sentences and using random in-domain context is better than using random out-of-domain context. We also found that for the given training/development/evaluation data combination the best result for \enja is achieved by using 2 context sentences, but for \jaen - by using 3.

For reference, we also trained baseline models on WMT20\footnote{http://www.statmt.org/wmt20/translation-task.html} data ($\sim$13M parallel sentences; WMT20 rows in Table \ref{tab:automatic-result-table}) and a mix of all data (WMT20+ rows). While these do outperform baselines trained only on the document-aligned data, the difference in automatic evaluation results is not too outstanding.

\section{Human Evaluation}

We perform human evaluation to compare the 0-context baseline and our highest scoring models (\textit{3-context} for \jaen and \textit{2-context} for \enja). 
Following the pairwise evaluation method from the WAT workshop \cite{nakazawa-etal-2019-overview}, we randomly sample 400 sentences from each translation direction and employ 5 evaluators to perform a blind comparative evaluation task by specifying if the translation is better or worse than the baseline (-1 or 1) or are they equal (0). Note that the evaluators had access to the context sentences so they can take the context into consideration for the evaluation, however, they had no access to the system names.
The final decision for a sentence is determined as a win if the sum of evaluations S $\geq$ 2, a loss if S $\leq$ -2, or a tie otherwise.
We calculate a pairwise score in range of -100 to 100  as follows: 
\[
    Pairwise = 100 \times \frac{W-L}{W+L+T},
\]
where a negative value favours the \textit{0-context} baseline and a positive value - the \textit{2/3-context} model.

Table \ref{tab:human-table} shows that in both directions models with actual context significantly outperform models with empty lines as context, even for \enja where the difference in BLEU scores was not significant.

We also calculated the Free-Marginal Kappa \cite{randolph2005free} values for the evaluations to measure inter-annotator agreement between evaluators. The results (\enja overall agreement - 64.90\%, Free-marginal kappa - 0.47; \jaen overall agreement - 67.65\%, Free-marginal kappa - 0.51) show intermediate to good agreement.
\begin{table}[t]
    \centering
    \begin{tabular}{|l|l|l|}
    \hline
              & \jaen & \enja      \\ \hline
    Wins      & 131   & 107        \\
    Losses    & 48    & 61         \\
    Ties      & 221   & 232        \\
    Score     & 20.75 ± 3.67 & 11.50 ± 3.50  \\
    Agreement & 67.65 & 64.90 \\
    Kappa     & 0.47  & 0.51       \\ \hline
    \end{tabular}
    \caption{Human evaluation results comparing wins, losses and ties for the \textit{2-context} \enja model and \textit{3-context} \jaen model against the \textit{0-context} models.}
    \label{tab:human-table}
\end{table}

% \begin{figure}[t]
%   \includegraphics[width=\linewidth]{figures/human-eval.pdf}
%   \caption{Human evaluation result visualisation.}
%   \label{fig:human-figure}
% \end{figure}

% We manually checked two subsets of \jaen evaluations - where all evaluators agreed that the baseline is better (S = -5; 7 in total) and where all agreed that the proposed model was better (S = 5; 37 in total). Out of the 

\section{Conclusion}

In this paper we explored how the data that is provided as context in the second source encoder of multi-source transformer models impacts the final translation quality. Firstly, we found that using only one previous sentence as context is not the optimal choice - two or three seem to be better, but this obviously depends on the data used, languages in question and translation direction. 

Another interesting finding is that the multi-source transformer model significantly outperformed the baseline transformer without any additional data at all. Our intuition is that this is due to the larger model architecture which sees the second empty source as noise and therefore learns clearer distinctions in the actual training data.

Lastly, we have shown that not all random data provided as previous context to multi-source transformer models has equal effect. Using in-domain random context led to 0.75 to 1.09 more BLEU than using out-of-domain random context, and both versions of random context were still slightly worse (0.08 - 0.31 BLEU) than the same corresponding models that used the correct context.

For future work we plan to perform similar experiments on different less explored language pairs, which is challenging due to the requirement of a decent amount of document aligned data, preferably with document boundaries. We would also be interested in probing the trained models and exploring what was learned by training on empty context lines.

% \section*{Acknowledgements}

% This work was supported by “Research and Development of Deep Learning Technology for Advanced Multilingual Speech Translation”, the Commissioned Research of National Institute of Information and Communications Technology (NICT), JAPAN.

\bibliography{anthology,custom}

\begin{thebibliography}{34}
\expandafter\ifx\csname natexlab\endcsname\relax\def\natexlab#1{#1}\fi

\bibitem[{Doddington(2002)}]{10.5555/1289189.1289273}
George Doddington. 2002.
\newblock Automatic evaluation of machine translation quality using n-gram
  co-occurrence statistics.
\newblock In \emph{Proceedings of the Second International Conference on Human
  Language Technology Research}, HLT '02, page 138–145, San Francisco, CA,
  USA. Morgan Kaufmann Publishers Inc.

\bibitem[{Hangyo et~al.(2014)Hangyo, Kawahara, and
  Kurohashi}]{Masatsugu-Hangyo2014}
Masatsugu Hangyo, Daisuke Kawahara, and Sadao Kurohashi. 2014.
\newblock \href {https://doi.org/10.5715/jnlp.21.213} {Building and analyzing a
  diverse document leads corpus annotated with semantic relations}.
\newblock \emph{Journal of Natural Language Processing}, 21(2):213--247.

\bibitem[{Jean et~al.(2017)Jean, Lauly, Firat, and Cho}]{jean2017does}
Sebastien Jean, Stanislas Lauly, Orhan Firat, and Kyunghyun Cho. 2017.
\newblock \href {http://arxiv.org/abs/1704.05135} {Does neural machine
  translation benefit from larger context?}

\bibitem[{Junczys-Dowmunt(2019)}]{junczys-dowmunt-2019-microsoft}
Marcin Junczys-Dowmunt. 2019.
\newblock \href {https://doi.org/10.18653/v1/W19-5321} {{M}icrosoft translator
  at {WMT} 2019: Towards large-scale document-level neural machine
  translation}.
\newblock In \emph{Proceedings of the Fourth Conference on Machine Translation
  (Volume 2: Shared Task Papers, Day 1)}, pages 225--233, Florence, Italy.
  Association for Computational Linguistics.

\bibitem[{Junczys-Dowmunt and
  Grundkiewicz(2018)}]{junczys-dowmunt-grundkiewicz-2018-ms}
Marcin Junczys-Dowmunt and Roman Grundkiewicz. 2018.
\newblock \href {https://doi.org/10.18653/v1/W18-6467} {{MS}-{UE}din submission
  to the {WMT}2018 {APE} shared task: Dual-source transformer for automatic
  post-editing}.
\newblock In \emph{Proceedings of the Third Conference on Machine Translation:
  Shared Task Papers}, pages 822--826, Belgium, Brussels. Association for
  Computational Linguistics.

\bibitem[{Junczys-Dowmunt et~al.(2018)Junczys-Dowmunt, Grundkiewicz, Dwojak,
  Hoang, Heafield, Neckermann, Seide, Germann, Aji, Bogoychev, Martins, and
  Birch}]{junczys-dowmunt-etal-2018-marian}
Marcin Junczys-Dowmunt, Roman Grundkiewicz, Tomasz Dwojak, Hieu Hoang, Kenneth
  Heafield, Tom Neckermann, Frank Seide, Ulrich Germann, Alham~Fikri Aji,
  Nikolay Bogoychev, Andr{\'e} F.~T. Martins, and Alexandra Birch. 2018.
\newblock \href {https://doi.org/10.18653/v1/P18-4020} {{M}arian: Fast neural
  machine translation in {C}++}.
\newblock In \emph{Proceedings of {ACL} 2018, System Demonstrations}, pages
  116--121, Melbourne, Australia. Association for Computational Linguistics.

\bibitem[{Jwalapuram et~al.(2020)Jwalapuram, Rychalska, Joty, and
  Basaj}]{jwalapuram2020benchmark}
Prathyusha Jwalapuram, Barbara Rychalska, Shafiq Joty, and Dominika Basaj.
  2020.
\newblock \href {https://arxiv.org/pdf/2004.14607.pdf} {Can your context-aware
  mt system pass the dip benchmark tests? : Evaluation benchmarks for discourse
  phenomena in machine translation}.
\newblock In \emph{Proceedings of the Workshop on Discourse in Machine
  Translation}. arXiv.

\bibitem[{Kim et~al.(2019)Kim, Tran, and Ney}]{kim-etal-2019-document}
Yunsu Kim, Duc~Thanh Tran, and Hermann Ney. 2019.
\newblock \href {https://doi.org/10.18653/v1/D19-6503} {When and why is
  document-level context useful in neural machine translation?}
\newblock In \emph{Proceedings of the Fourth Workshop on Discourse in Machine
  Translation (DiscoMT 2019)}, pages 24--34, Hong Kong, China. Association for
  Computational Linguistics.

\bibitem[{Koehn(2004)}]{koehn-2004-statistical}
Philipp Koehn. 2004.
\newblock \href {https://www.aclweb.org/anthology/W04-3250} {Statistical
  significance tests for machine translation evaluation}.
\newblock In \emph{Proceedings of the 2004 Conference on Empirical Methods in
  Natural Language Processing}, pages 388--395, Barcelona, Spain. Association
  for Computational Linguistics.

\bibitem[{Kudo(2006)}]{kudo2006mecab}
Taku Kudo. 2006.
\newblock Mecab: Yet another part-of-speech and morphological analyzer.
\newblock \emph{http://mecab. sourceforge. jp}.

\bibitem[{Kudo and Richardson(2018)}]{kudo-richardson-2018-sentencepiece}
Taku Kudo and John Richardson. 2018.
\newblock \href {https://doi.org/10.18653/v1/D18-2012} {{S}entence{P}iece: A
  simple and language independent subword tokenizer and detokenizer for neural
  text processing}.
\newblock In \emph{Proceedings of the 2018 Conference on Empirical Methods in
  Natural Language Processing: System Demonstrations}, pages 66--71, Brussels,
  Belgium. Association for Computational Linguistics.

\bibitem[{Li et~al.(2020)Li, Liu, Wang, Jiang, Xiao, Zhu, Liu, and
  Li}]{li-etal-2020-multi-encoder}
Bei Li, Hui Liu, Ziyang Wang, Yufan Jiang, Tong Xiao, Jingbo Zhu, Tongran Liu,
  and Changliang Li. 2020.
\newblock \href {https://doi.org/10.18653/v1/2020.acl-main.322} {Does
  multi-encoder help? a case study on context-aware neural machine
  translation}.
\newblock In \emph{Proceedings of the 58th Annual Meeting of the Association
  for Computational Linguistics}, pages 3512--3518, Online. Association for
  Computational Linguistics.

\bibitem[{Lopes et~al.(2020)Lopes, Farajian, Bawden, Zhang, and
  Martins}]{lopes-etal-2020-document}
Ant{\'o}nio Lopes, M.~Amin Farajian, Rachel Bawden, Michael Zhang, and
  Andr{\'e} F.~T. Martins. 2020.
\newblock \href {https://www.aclweb.org/anthology/2020.eamt-1.24}
  {Document-level neural {MT}: A systematic comparison}.
\newblock In \emph{Proceedings of the 22nd Annual Conference of the European
  Association for Machine Translation}, pages 225--234, Lisboa, Portugal.
  European Association for Machine Translation.

\bibitem[{Macé and Servan(2019)}]{mace_valentin_2019_3525020}
Valentin Macé and Christophe Servan. 2019.
\newblock \href {https://doi.org/10.5281/zenodo.3525020} {{Using Whole Document
  Context in Neural Machine Translation}}.
\newblock In \emph{16th International Workshop on Spoken Language Translation
  2019}. Zenodo.

\bibitem[{Maruf et~al.(2019)Maruf, Martins, and
  Haffari}]{maruf-etal-2019-selective}
Sameen Maruf, Andr{\'e} F.~T. Martins, and Gholamreza Haffari. 2019.
\newblock \href {https://doi.org/10.18653/v1/N19-1313} {Selective attention for
  context-aware neural machine translation}.
\newblock In \emph{Proceedings of the 2019 Conference of the North {A}merican
  Chapter of the Association for Computational Linguistics: Human Language
  Technologies, Volume 1 (Long and Short Papers)}, pages 3092--3102,
  Minneapolis, Minnesota. Association for Computational Linguistics.

\bibitem[{Miculicich et~al.(2018)Miculicich, Ram, Pappas, and
  Henderson}]{miculicich-etal-2018-document}
Lesly Miculicich, Dhananjay Ram, Nikolaos Pappas, and James Henderson. 2018.
\newblock \href {https://doi.org/10.18653/v1/D18-1325} {Document-level neural
  machine translation with hierarchical attention networks}.
\newblock In \emph{Proceedings of the 2018 Conference on Empirical Methods in
  Natural Language Processing}, pages 2947--2954, Brussels, Belgium.
  Association for Computational Linguistics.

\bibitem[{Morishita et~al.(2020)Morishita, Suzuki, and
  Nagata}]{morishita-etal-2020-jparacrawl}
Makoto Morishita, Jun Suzuki, and Masaaki Nagata. 2020.
\newblock \href {https://www.aclweb.org/anthology/2020.lrec-1.443}
  {{JP}ara{C}rawl: A large scale web-based {E}nglish-{J}apanese parallel
  corpus}.
\newblock In \emph{Proceedings of the 12th Language Resources and Evaluation
  Conference}, pages 3603--3609, Marseille, France. European Language Resources
  Association.

\bibitem[{Nakazawa et~al.(2019)Nakazawa, Doi, Higashiyama, Ding, Dabre, Mino,
  Goto, Pa, Kunchukuttan, Oda, Parida, Bojar, and
  Kurohashi}]{nakazawa-etal-2019-overview}
Toshiaki Nakazawa, Nobushige Doi, Shohei Higashiyama, Chenchen Ding, Raj Dabre,
  Hideya Mino, Isao Goto, Win~Pa Pa, Anoop Kunchukuttan, Yusuke Oda,
  Shantipriya Parida, Ond{\v{r}}ej Bojar, and Sadao Kurohashi. 2019.
\newblock \href {https://doi.org/10.18653/v1/D19-5201} {Overview of the 6th
  workshop on {A}sian translation}.
\newblock In \emph{Proceedings of the 6th Workshop on Asian Translation}, pages
  1--35, Hong Kong, China. Association for Computational Linguistics.

\bibitem[{Pal et~al.(2018)Pal, Herbig, Kr{\"u}ger, and van
  Genabith}]{pal-etal-2018-transformer}
Santanu Pal, Nico Herbig, Antonio Kr{\"u}ger, and Josef van Genabith. 2018.
\newblock \href {https://doi.org/10.18653/v1/W18-6468} {A transformer-based
  multi-source automatic post-editing system}.
\newblock In \emph{Proceedings of the Third Conference on Machine Translation:
  Shared Task Papers}, pages 827--835, Belgium, Brussels. Association for
  Computational Linguistics.

\bibitem[{Papineni et~al.(2002)Papineni, Roukos, Ward, and
  Zhu}]{papineni-etal-2002-bleu}
Kishore Papineni, Salim Roukos, Todd Ward, and Wei-Jing Zhu. 2002.
\newblock \href {https://doi.org/10.3115/1073083.1073135} {{B}leu: a method for
  automatic evaluation of machine translation}.
\newblock In \emph{Proceedings of the 40th Annual Meeting of the Association
  for Computational Linguistics}, pages 311--318, Philadelphia, Pennsylvania,
  USA. Association for Computational Linguistics.

\bibitem[{Popovi{\'c}(2015)}]{popovic-2015-chrf}
Maja Popovi{\'c}. 2015.
\newblock \href {https://doi.org/10.18653/v1/W15-3049} {chr{F}: character
  n-gram {F}-score for automatic {MT} evaluation}.
\newblock In \emph{Proceedings of the Tenth Workshop on Statistical Machine
  Translation}, pages 392--395, Lisbon, Portugal. Association for Computational
  Linguistics.

\bibitem[{Post(2018)}]{post-2018-call}
Matt Post. 2018.
\newblock \href {https://doi.org/10.18653/v1/W18-6319} {A call for clarity in
  reporting {BLEU} scores}.
\newblock In \emph{Proceedings of the Third Conference on Machine Translation:
  Research Papers}, pages 186--191, Brussels, Belgium. Association for
  Computational Linguistics.

\bibitem[{Randolph(2005)}]{randolph2005free}
Justus~J Randolph. 2005.
\newblock Free-marginal multirater kappa (multirater $\kappa$free): An
  alternative to fleiss’ fixed-marginal multirater kappa.
\newblock In \emph{Presented at the Joensuu Learning and Instruction
  Symposium}, volume 2005.

\bibitem[{Rikters et~al.(2020)Rikters, Ri, Li, and
  Nakazawa}]{rikters-EtAl:2020:WMT}
Matīss Rikters, Ryokan Ri, Tong Li, and Toshiaki Nakazawa. 2020.
\newblock \href {https://www.aclweb.org/anthology/2020.wmt-1.74}
  {Document-aligned japanese-english conversation parallel corpus}.
\newblock In \emph{Proceedings of the Fifth Conference on Machine Translation},
  pages 637--643, Online. Association for Computational Linguistics.

\bibitem[{Shin and Lee(2018)}]{shin-lee-2018-multi}
Jaehun Shin and Jong-Hyeok Lee. 2018.
\newblock \href {https://doi.org/10.18653/v1/W18-6470} {Multi-encoder
  transformer network for automatic post-editing}.
\newblock In \emph{Proceedings of the Third Conference on Machine Translation:
  Shared Task Papers}, pages 840--845, Belgium, Brussels. Association for
  Computational Linguistics.

\bibitem[{Stojanovski and Fraser(2020)}]{stojanovski2020addressing}
Dario Stojanovski and Alexander Fraser. 2020.
\newblock \href {http://arxiv.org/abs/2004.14927} {Addressing zero-resource
  domains using document-level context in neural machine translation}.

\bibitem[{Tiedemann and Scherrer(2017)}]{tiedemann-scherrer-2017-neural}
J{\"o}rg Tiedemann and Yves Scherrer. 2017.
\newblock \href {https://doi.org/10.18653/v1/W17-4811} {Neural machine
  translation with extended context}.
\newblock In \emph{Proceedings of the Third Workshop on Discourse in Machine
  Translation}, pages 82--92, Copenhagen, Denmark. Association for
  Computational Linguistics.

\bibitem[{Tu et~al.(2018)Tu, Liu, Shi, and Zhang}]{tu-etal-2018-learning}
Zhaopeng Tu, Yang Liu, Shuming Shi, and Tong Zhang. 2018.
\newblock \href {https://doi.org/10.1162/tacl_a_00029} {Learning to remember
  translation history with a continuous cache}.
\newblock \emph{Transactions of the Association for Computational Linguistics},
  6:407--420.

\bibitem[{Vaswani et~al.(2017)Vaswani, Shazeer, Parmar, Uszkoreit, Jones,
  Gomez, Kaiser, and Polosukhin}]{NIPS2017_3f5ee243}
Ashish Vaswani, Noam Shazeer, Niki Parmar, Jakob Uszkoreit, Llion Jones,
  Aidan~N Gomez, \L~ukasz Kaiser, and Illia Polosukhin. 2017.
\newblock \href
  {https://proceedings.neurips.cc/paper/2017/file/3f5ee243547dee91fbd053c1c4a845aa-Paper.pdf}
  {Attention is all you need}.
\newblock In \emph{Advances in Neural Information Processing Systems},
  volume~30, pages 5998--6008. Curran Associates, Inc.

\bibitem[{Voita et~al.(2018)Voita, Serdyukov, Sennrich, and
  Titov}]{voita-etal-2018-context}
Elena Voita, Pavel Serdyukov, Rico Sennrich, and Ivan Titov. 2018.
\newblock \href {https://doi.org/10.18653/v1/P18-1117} {Context-aware neural
  machine translation learns anaphora resolution}.
\newblock In \emph{Proceedings of the 56th Annual Meeting of the Association
  for Computational Linguistics (Volume 1: Long Papers)}, pages 1264--1274,
  Melbourne, Australia. Association for Computational Linguistics.

\bibitem[{Yao and Wan(2020)}]{yao-wan-2020-multimodal}
Shaowei Yao and Xiaojun Wan. 2020.
\newblock \href {https://doi.org/10.18653/v1/2020.acl-main.400} {Multimodal
  transformer for multimodal machine translation}.
\newblock In \emph{Proceedings of the 58th Annual Meeting of the Association
  for Computational Linguistics}, pages 4346--4350, Online. Association for
  Computational Linguistics.

\bibitem[{Zhang et~al.(2018)Zhang, Luan, Sun, Zhai, Xu, Zhang, and
  Liu}]{zhang-etal-2018-improving}
Jiacheng Zhang, Huanbo Luan, Maosong Sun, Feifei Zhai, Jingfang Xu, Min Zhang,
  and Yang Liu. 2018.
\newblock \href {https://doi.org/10.18653/v1/D18-1049} {Improving the
  transformer translation model with document-level context}.
\newblock In \emph{Proceedings of the 2018 Conference on Empirical Methods in
  Natural Language Processing}, pages 533--542, Brussels, Belgium. Association
  for Computational Linguistics.

\bibitem[{Zhou et~al.(2020)Zhou, Yılmaz, Long, Li, and Li}]{Zhou2020}
Xinyuan Zhou, Emre Yılmaz, Yanhua Long, Yijie Li, and Haizhou Li. 2020.
\newblock \href {https://doi.org/10.21437/Interspeech.2020-2488}
  {{Multi-Encoder-Decoder Transformer for Code-Switching Speech Recognition}}.
\newblock In \emph{Proc. Interspeech 2020}, pages 1042--1046.

\bibitem[{Zoph and Knight(2016)}]{zoph-knight-2016-multi}
Barret Zoph and Kevin Knight. 2016.
\newblock \href {https://doi.org/10.18653/v1/N16-1004} {Multi-source neural
  translation}.
\newblock In \emph{Proceedings of the 2016 Conference of the North {A}merican
  Chapter of the Association for Computational Linguistics: Human Language
  Technologies}, pages 30--34, San Diego, California. Association for
  Computational Linguistics.

\end{thebibliography}
\bibliographystyle{acl_natbib}

% \appendix

% \section{Example Appendix}
% \label{sec:appendix}

% This is an appendix.

\end{document}